\documentclass{article} % For LaTeX2e
\usepackage{iclr2026_conference,times}

\usepackage{amsmath,amsfonts,bm}

\def\eqref#1{equation~\ref{#1}}
\def\1{\bm{1}}

\DeclareMathAlphabet{\mathsfit}{\encodingdefault}{\sfdefault}{m}{sl}
\SetMathAlphabet{\mathsfit}{bold}{\encodingdefault}{\sfdefault}{bx}{n}

\usepackage{hyperref}
\usepackage{url}
\usepackage{booktabs} % For formal tables
\usepackage{multirow}
\usepackage{graphicx}
\usepackage{makecell}
\usepackage{wrapfig}
\usepackage{wrapfig}
\usepackage{xcolor}

\title{GeoSDF: Plane Geometry Diagram Synthesis via Signed Distance Field}

\author{
\textbf{Chengrui Zhang}$^{1,}$\thanks{Both authors contributed equally to this paper. $^\dagger$Corresponding author.}, 
\textbf{Maizhen Ning}$^{1,}$\footnotemark[1], 
\textbf{Tianyi Liu}$^{1}$, 
\textbf{Zihao Zhou}$^{1}$, \\
\textbf{Jie Sun}$^{1, \dagger}$, 
\textbf{Qiufeng Wang}$^{1, \dagger}$, 
\textbf{Kaizhu Huang}$^{2}$ \\
\textsuperscript{1} Xi'an Jiaotong-Liverpool University, Suzhou, China, 
\textsuperscript{2} Duke Kunshan University, Kunshan, China \\
\texttt{\small \{Jie.Sun, Qiufeng.Wang\}@xjtlu.edu.cn, }
\texttt{\small Kaizhu.Huang@dukekunshan.edu.cn}
}
\iclrfinalcopy % Uncomment for camera-ready version, but NOT for submission.
\begin{document}

\maketitle

\begin{abstract}
Plane Geometry Diagram Synthesis has been a crucial task in computer graphics, with applications ranging from educational tools to AI-driven mathematical reasoning. Traditionally, we rely on manual tools (e.g., Matplotlib and GeoGebra) to generate precise diagrams, but this usually requires huge, complicated calculations. Recently, researchers start to work on model-based methods (e.g., Stable Diffusion and GPT5) to automatically generate diagrams, saving operational cost but usually suffering from limited realism and insufficient accuracy. In this paper, we propose a novel framework GeoSDF, to automatically generate diagrams efficiently and accurately with Signed Distance Field (SDF). Specifically, we first represent geometric elements (e.g., points, segments, and circles) in the SDF, then construct a series of constraint functions to represent geometric relationships. Next, we optimize those constructed constraint functions to get an optimized field of both elements and constraints. Finally, by rendering the optimized field, we can obtain the synthesized diagram. In our GeoSDF, we define a symbolic language to represent geometric elements and constraints, and our synthesized geometry diagrams can be self-verified in the SDF, ensuring both mathematical accuracy and visual plausibility. In experiments, through both qualitative and quantitative analysis, GeoSDF synthesized both normal high-school level and IMO-level geometry diagrams. We achieve 88.67\% synthesis accuracy by human evaluation in the IMO problem set. Furthermore, we obtain a very high accuracy of solving geometry problems (over 95\% while the current SOTA accuracy is around 75\%) by leveraging our self-verification property. All of these demonstrate the advantage of GeoSDF, paving the way for more sophisticated, accurate, and flexible generation of geometric diagrams for a wide array of applications. 
% The accompanying code, datasets, and all synthesized outputs are being released to benefit the research community.
\end{abstract}

\maketitle

\section{Introduction}
\begin{figure*}[ht!]
    \centering
    \includegraphics[width=1\linewidth]{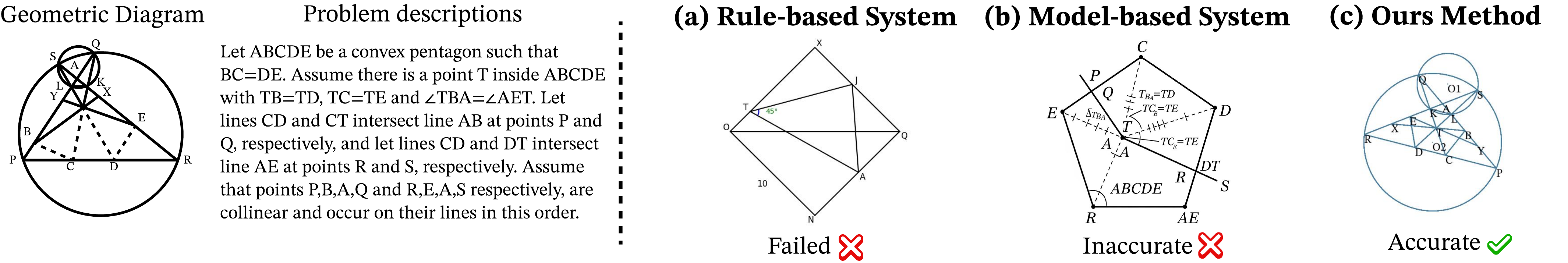}
    \caption{Comparison of geometric diagram synthesis methods for IMO 2022, Problem 4. (a) Rule-based systems, such as R-COT, rely on predefined templates and fail to produce diagrams when faced with non-predefined geometric constraints. (b) General model-based systems, including Gemini, DALL-E 3, and GPT-4o, often lack the precision for accurate mathematical synthesis, resulting in geometrically inconsistent figures. (c) In contrast, our proposed method accurately constructs the diagram by interpreting the fundamental geometric elements and constraints directly from the problem description.
    }
    \label{fig:banner}
\end{figure*}
% reference: https://www.geogebra.org/classic/s6XBVT5K https://www.geogebra.org/m/W8yej6Xy

The study of geometry problems is fundamental across numerous disciplines, serving as a cornerstone for advancements in computer graphics~\cite{ye2020penrose}, computational geometry~\cite{dorst2002geometric}, and even theoretical mathematics~\cite{AlphaGeometryTrinh2024}. The emergence of large language models (LLMs) has introduced new approaches for addressing Plane Geometry Problems (PGPs)~\cite{shi2024continual} by understanding both textual problem descriptions and their corresponding geometric diagrams~\cite{g-llava}.
% ,geogpt4v,geox,xu2024geo,mathvista,mathverse,geoeval}. 
Despite significant efforts to enhance the reasoning capabilities of LLMs, the performance of models remains limited by the scarcity of high-quality datasets containing annotated plane geometry diagrams~\cite{zhang2023formalgeo}. 

There are several techniques designed to improve the construction of geometric diagrams.
A straightforward approach is manual construction, such as Matplotlib~\cite{Hunter:2007} and GeoGebra~\cite{geogebra}.
However, this process is often cumbersome and time-consuming, as it requires explicit specification of element positions and offers limited flexibility for handling complex configurations.
% manual system
% Recent manual system advancements~\cite{Hunter:2007, geogebra, kazemi2024geomverse} have led to the development of tools specifically designed to enhance the quality of geometric diagrams. However, as illustrated in Figure~\ref{fig:banner},  manual construction using computer tools such as Matplotlib~\cite{Hunter:2007} and GeoGebra~\cite{geogebra} typically requires the explicit specification of element positions. This process is often inflexible and laborious, with a large amount of calculation, as it necessitates adapting to high-level geometric relationships or ensuring the satisfaction of intricate constraints. 
% rule-based
To accelerate diagram generation, rule-based approaches have been proposed~\cite{kazemi2024geomverse}, which employ programmatic techniques to automate construction. Nonetheless, these methods rely heavily on predefined geometric primitives and fixed assembly rules, which pose limitations when generating diagrams with complex or irregular configurations. 
% diffusion
The remarkable capabilities of model-based methods like Stable Diffusion~\cite{rombach2021highresolution}, GPT5, and DALL$\cdot$E~3~\cite{dalle3} in producing realistic natural images offer an important insight: it raises the question of whether similar approaches can be leveraged to generate geometric diagrams. Unfortunately, as illustrated in Figure~\ref{fig:banner}, these methods struggle with the deterministic rules and mathematical validity (\textit{e.g.,} exact coordinates, relations), resulting in inaccurate diagrams~\cite{controlnet, zhang2023formalgeo}.

In this paper, we propose a novel framework to automatically synthesize precise \textbf{Geo}metry diagrams by leveraging \textbf{S}igned \textbf{D}istance \textbf{F}ield (SDF), termed as \textbf{GeoSDF}. 
% In the theory of SDF, it defines a scalar field that assigns to each point in space a signed distance to the nearest surface of a geometric object~\cite{yariv2020nerfsdf}, thus being differentiable to be optimization. In our proposed GeoSDF, the textual problem description is first parsed into a set of mathematically well-defined components, which we refer to as \textbf{elements}. These elements are then governed by a set of \textbf{geometric constraints} derived from the problem's underlying mathematical relationships, such as perpendicularity, parallelism, or incidence. Unlike previous methods, our GeoSDF allows these relationships to be quantitatively expressed—for instance, through the distance from a point to a line—enabling precise and verifiable encoding of geometric structure. These constraints are then incorporated into a loss function, effectively translating the mathematical conditions into an \textbf{optimization objective}. By leveraging gradient descent, GeoSDF iteratively refines the positions of elements in the diagram to satisfy complex constraints, allowing for the generation of accurate and mathematically valid geometric figures. This formulation supports both quantifiable relations and the derivation of exact solutions, addressing key limitations of existing diagram generation methods. 
In the first step, problem descriptions are parsed into a structured symbolic representation consisting of \textbf{elements} (e.g., points, lines, circles) as well as the relationship among them, which are termed as \textbf{constraints}.
We then represent each geometric element as a signed distance field (SDF), a continuous and differentiable function that compactly encodes its shape and position.
Geometric constraints are expressed as differentiable functions over these fields, allowing us to formulate a global loss that enforces all constraints simultaneously.
To prevent degenerate configurations such as overlapping elements, we additionally introduce a crowd regularization term that encourages well-spaced solutions.
The resulting differentiable objective can be optimized using standard gradient-based methods. Finally, the optimized geometric elements are rendered into clear, human-readable diagrams, providing precise diagrams that faithfully reflect the mathematical relationships.

% 我们在xx数据集上验证了我们的工具，很直接的一点使我们比前面的东西都好，因为他们画不出难得题
% 在xx数据集上 我们得效果是 数值是xxx xxx，这种高置信度说我们我们的工具使可靠的
% 之前的好用改的工具用他们自己得语言，我们是直接用llm转换得symbolic language,  这使我们工具更加简单使用。 同时，我们在优化得过程中，我们可以按batch进行操作，也就是同时操作N涨图，这使我我们得作图工作更加高效。
% 我们同时release了这两个数据集的文章的直接转换的成果。

Previous works~\cite{ye2020penrose} adopted a hard learning curve of domain-specific languages; in contrast, we leverage Large Language Models (LLMs) to translate natural language directly into a symbolic format.
Based on simple use, our framework has the state-of-the-art performance on downstream tasks. We demonstrate the effectiveness on a range of challenging datasets, from the high-school level to an IMO-level benchmark. Our method achieves a problem-solving accuracy of 95.9\%, a high-confidence result that confirms the reliability of our tool and surpasses the previous best by over 20 percentage points on the GeoQA dataset. And, our method enables the successful synthesis of diagrams for IMO-level problems where prior methods fall short. Moreover, our optimization process operates in batches, enabling the highly efficient, parallel synthesis of multiple diagrams. 

The contributions of this paper are summarized as follows:
\begin{itemize}
\item We introduce GeoSDF, a novel and accurate framework for synthesizing plane geometry diagrams by optimizing SDF representation against symbolic mathematical constraints.
\item We design a more accessible user workflow where natural language is translated by an LLM into a symbolic representation, eliminating the need for specialized languages required by previous tools.
\item We achieve high efficiency through a batch-enabled optimization process, allowing for the simultaneous generation of multiple complex diagrams.
% \item We construct and release two new, large-scale datasets of geometric problems, including an IMO-level dataset, to specifically address the synthesis of complex and challenging diagrams.
\item To demonstrate our method's ability to generate high quantity and quality outputs, we augmented the existing data by synthesizing 32 diagrams with diversity for each problem description in FormalGeo7k dataset, finally forming a large size of dataset with 224k samples.
\item We achieve SOTA results on the GeoQA benchmark with 95.9\% accuracy, demonstrating that high-quality diagram synthesis directly translates to superior geometric problem-solving capabilities.
\end{itemize}

\section{Related Work}%\Background}

\subsection{Plane Geometry Problem Solving}
Early approaches to plane geometry problem (PGP) solving relied on manually defined rules applied to small datasets~\cite{rw_gp_geos_one,rw_gp_geos_two}, resulting in poor generalization. Neural network-based models, such as NGS~\cite{geoqa} and DPE-NGS~\cite{geoqa}, introduced visual question answering and specialized program generation, but still struggle with coarse-grained diagram understanding~\cite{sca-gps}. Symbolic reasoning methods, including Inter-GPS~\cite{inter-gps} and FormalGeo~\cite{zhang2023formalgeo}, use complex rule-based systems to interpret formal languages parsed from diagrams and text. However, their performance is constrained by limited datasets and parameter sizes, and they do not produce natural language solutions. Recent advances in multi-modal large language models (MLLMs), such as G-LLaVA~\cite{g-llava} and Geo-LLaVA~\cite{geo-llava}, fine-tune based models~\cite{llava1.5} to generate natural language solutions. These approaches benefit from data augmentation techniques (e.g., Geo170K via GPT-3.5~\cite{g-llava}), but often prioritize textual information over the diversity and complexity of geometry diagrams~\cite{mathpuma}. While methods like GeoX~\cite{geox} and R-CoT~\cite{r-cot} have made progress in diagram understanding, generating accurate, diverse, and controllable geometric diagrams remains a fundamental challenge for advancing PGP solving.

\subsection{Plane Geometry Diagram Synthesis}
Plane geometry diagram synthesis is critical for evaluating problem-solving systems but is underexplored due to complexity~\cite{rombach2022high}. Current rule-based methods, including GeomVerse~\cite{kazemi2024geomverse} and MAVIS~\cite{mavis}, stitch predefined shapes along edges. However, this limits them from generating diagrams with inscribed relationships (e.g., a triangle in a circle). Despite R-CoT's~\cite {r-cot} advancements with inscribed elements, existing approaches remain constrained by predefined shapes and rules, preventing synthesis from textual requirements. 
Another approach is Penrose~\cite{ye2020penrose}, which synthesizes diagrams from a domain-specific language by solving a constrained numerical optimization problem. However, for problems in Euclidean geometry that demand high mathematical fidelity, its reliance on a general-purpose constraint solver can be less ideal. Our work differs fundamentally in its representation: GeoSDF leverages Signed Distance Fields to directly embed core geometric properties—like distance, angle, and relations—into the loss functions. This specialized approach results in a more stable and direct convergence to mathematically precise configurations.
% Our proposed GeoSDF overcomes this by synthesizing flexible plane geometry diagrams directly from given text constraints. 

\section{Methodology}\label{methodology}
\begin{figure*}[ht]
    \centering
    \includegraphics[width=1\linewidth]{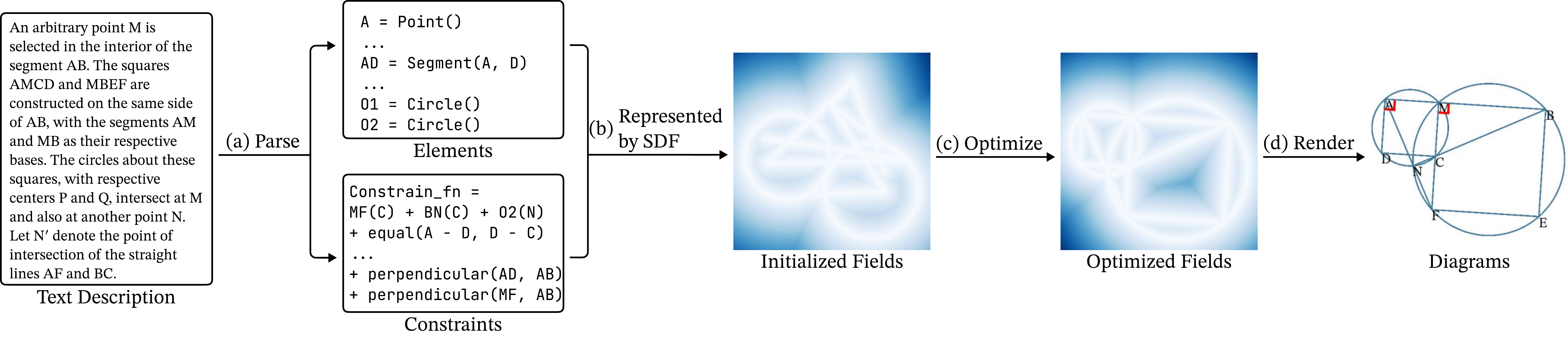}
    \caption{The GeoSDF Pipeline with an example IMO diagram synthesis process. (a) Parsing text description in natural language to geometric Elements and Constraints in symbolic language. (b) The elements and constraints are represented using SDF representations. (c) The optimization process from the initialized field to satisfy all the specified constraints forms the optimized Fields. (d) The final optimized fields are rendered to Diagrams that visualize the geometric problem.}
    \label{fig:pipeline}
\end{figure*}

The GeoSDF framework is developed for synthesizing plane geometry diagrams and consists of four steps, as illustrated in Fig.~\ref{fig:pipeline}. Each step is described in detail in the following sections.

\subsection{Parsing Natural Language into Symbolic Language}

\begin{table*}[h!]
    \centering
    \caption{Predefined common constraints and loss functions in GeoSDF. In our framework, GeoSDF, we consider these geometric constraints directly as the loss functions to quantify the deviation from the desired configuration. The mathematical details are in Appendix~\ref{app:constraints}. Users could easily extend the extra loss function to satisfy other customized constraints.}
    \resizebox{1.0 \textwidth}{!}{
        \begin{tabular}{llll}
            \toprule
            \textbf{Category}           & \textbf{Constraint} & \textbf{Loss Function}            & \textbf{Description}                                                     \\
            \midrule
            \multirow{2}{*}{Incidence} & Equality            & $\text{Equal}(A, B)$          & Two geometric elements $A, B$ have the same value.                   \\
                                                   & Less                & $\text{Less}(A, B)$           & One numerical value is less than or equal to another.                    \\
            \midrule
            \multirow{3}{*}{Metric}    & Distance            & $\text{Equal}(A(B),v)$            & A specific distance $v$ between two geometric elements $A~\text{to}~B$.       \\
                                                   & Angle               & $\text{Angle}(A,B,C)$       & The angle formed by three points $A,B,C$                           \\
                                                   & Area                & $\text{Area}(A, ..., E)$      & A specific area for a polygon defined by a sequence of points.           \\
                                                   % & Crowd Penalty       & $\text{Crowd}(x_1, ..., x_n)$     & Encourages geometric elements (typically points) to be spaced apart.     \\
            \midrule
            \multirow{4}{*}{Spatial} & Parallelism & $\text{Parallel}(AB, CD)$ & Two lines or segments are parallel to each other. \\                                  & Perpendicularity    & $\text{Perpendicular}(AB,CD)$ & Two line or segments are perpendicular to each other.                                         \\
                                                   & Order               & $\text{Order}(A, ..., E)$     & Counterclockwise order of points $A,...,E$.                          \\
                                                   & On                  & $AB(C)$                            & A specified point $C$ lies on a given segment or circle $AB$.             \\
                                                   & Inside              & $\text{Inside}(p, A, ..., E)$ & A point $p$ if it lies outside a convex polygon defined by other points $A$ to $E$. \\
            \midrule
            Topological                & Convexity           & $\text{Convex}(A, ..., E)$    & A polygon formed by a sequence of points $A,...,E$ is convex.        \\
            \bottomrule
        \end{tabular}
    }

    \label{tab:constraints}
\end{table*}

The first parsing step of the GeoSDF framework converts geometric natural language statements into a structured symbolic language representation in an end-to-end manner. 
As illustrated in Fig.~\ref{fig:pipeline}, the symbolic language consists of \textbf{Elements}, which denote fundamental geometric objects such as points, lines, or segments, and \textbf{Constraints}, which describe relationships or properties among these objects. Following previous works~\cite{Computational_Geometry}, we consider four types of constraints: (1) incidence (e.g., a point lies on a circle), (2) metric (e.g., distance between two points equals a given value), (3) spatial (e.g., two lines are parallel), and (4) topological (e.g., convexity is preserved), concluded in Table~\ref{tab:constraints}.

For instance, consider a case where two line segments are defined by points \texttt{A, B} and \texttt{C, D}, respectively, and the two segments are parallel. After processing through this step, the structured symbolic representation encodes the elements as \texttt{AB = SEGMENT(A, B)} and \texttt{CD = SEGMENT(C, D)}, while the constraint \texttt{Parallel(AB, CD)} captures the orientation relationship between the two segments.
We fine-tune a QWen2.5-7B-Instruct model~\cite{qwen25technicalreport} on the training data in FormalGeo-7K~\cite{zhang2023formalgeo} to obtain the model for the parsing process.

\subsection{Representation by SDF}

We represent a \textbf{geometric diagram}, composed of the aforementioned \textbf{elements and constraints}, as a collection of signed distance fields (SDFs). An SDF is a continuous and differentiable scalar field that assigns to each point $(x,y) \in \mathbb{R}^2$ its shortest distance to a given geometric object. More details about SDF can be found in Appendix~\ref{app:sdf}.

Each \textbf{element} in the diagram is represented by its own SDF, which is fully determined by the element’s continuous parameters (e.g., coordinates for points, radii for circles, lengths for segments). This representation captures the element’s shape and position, providing a differentiable field that can later be optimized.

\textbf{Constraints} that encode geometric relationships between elements are expressed as differentiable functions of  element parameters. They do not introduce new fields, rather, they specify conditions under which the combined element fields form a valid diagram. Each constraint thus acts as an optimization objective over the corresponding SDFs.

Finally, the complete diagram SDF is obtained by composing the individual element fields into a unified representation. 
% As illustrated in Fig.~\ref{fig:pipeline}, the initialized field is progressively refined into a consistent diagram where all constraints are satisfied.
As illustrated in Fig.~\ref{fig:pipeline}, we begin by initializing a random field for each element. In the visualization, lighter colors indicate smaller SDF values, corresponding to points closer to the element. By sampling the field, we can observe the underlying structure of the initialized diagram.
During optimization, the element parameters are adjusted to satisfy all constraints simultaneously, yielding a valid geometric configuration. We show a complete example in Appendix~\ref{app:Synthesize_Process}.

\subsection{Optimizing SDF Field} 
% let $E = { e_1, \dots, e_N }$ denote the set of elements and $C = \bigcup_{j=1}^N C_j$ the set of all constraints, where $C_j = { c_{j,k} }$ is the set of constraints involving element $e_j$. A valid configuration is achieved when all constraints in $C$ are satisfied simultaneously. 
% Since both elements and constraints are represented in differentiable form, we can optimize the diagram field using gradient-based methods.
To clarify the optimization step, the \textbf{diagram} shows the complete set of $N$ geometric elements, $E = \{ e_1, e_2, \dots, e_N \}$.
Each element $e_j \in E$ is associated with its own set of differentiable constraints
$C_j = \{ c_{j,k} \mid k = 1, 2, \dots, M_j \}$,
where $M_j$ is the number of constraints associated with element $e_j$, $k$ explicitly indexes the constraints for element $e_j$, so that $c_{j,k}$ denotes the $k$-th constraint in $C_j$. 
% These constraints encode geometric relations involving $e_j$, 
% and different elements may have different numbers or types of associated constraints. 
% In the initialization stage

The complete set of constraints across all elements is then $C = \bigcup_{j=1}^{N} C_j$.
A valid geometric configuration is obtained when all constraints in $C$ are satisfied simultaneously.  
To enforce all geometric constraints, we have the loss function:
% we define a differentiable loss function over the complete set of elements $E = \{ e_1, \dots, e_N \}$ and their associated constraints $C = \bigcup_{j=1}^N C_j$:
\begin{equation}
L_{constraints}(E, C) = \sum_{j=1}^{N} \sum_{c_{j,k} \in C_j} c_{j,k}(E),
\end{equation}
where each $c_{j,k}$ is transformed into a non-negative differentiable function that evaluates to zero when the corresponding constraint is satisfied, ensuring the loss is bounded below by zero and suitable for gradient-based optimization.

To prevent elements from collapsing into overlapping or degenerate configurations, which could otherwise cause the optimization to fail, we introduce a crowd regularization term: 
\begin{equation}
\label{eqn:reg}
L_{crowd}(E) = \sum_{1 \le i < j \le N} \left[ \max\left(0, \tau_r - \| x_i - x_j \| \right) \right]^2,
\end{equation}
where $x_i$ and $x_j$ denote the position parameters of elements $e_i$ and $e_j$, 
and the hyper-parameter $\tau_r$ specifies a distance between any pair of elements. 

The differentiable term can be optimized jointly with the SDF-based constraint losses.
The total loss over all elements and constraints is then defined as:
\begin{equation}
\label{eqn:total_loss}
L_{total} = L_{constraints}(E, C) + \lambda L_{crowd}(E),
\end{equation}
where $\lambda$ balances the relative importance of the regularization.  
This differentiable loss is minimized using standard gradient-based optimization, yielding a configuration $E^*$ in which constraints are approximately satisfied and elements remain well-spaced.  
Further visualization of the optimization procedure is provided in Appendix~\ref{app:optimization_process_visualization}.

% 为啥要加优化过程的那张图？能说明什么？

\subsection{Boundary Extraction and Visualization} 
The optimized configuration \(E^*\) specifies the final shapes of all geometric elements. Each element's geometric parameters (e.g., center and radius of a circle) uniquely determine its boundary. Therefore, the boundaries can be directly computed from the elements representation \(E^*\). 
% \red{A threshold parameter \(\tau_t\) is used to control the rendered boundary thickness, with larger values producing thicker lines and smaller values producing finer details. }  
% 这个粗的细的有必要说吗？？？？？
% 怎么前面也tau后面也tau??????
Detailed boundary extraction procedures are provided in Appendix~\ref{app:boundary_threshold}.
% 这里想给什么看？？？还没说明白？？？？？

% \begin{wrapfigure}{R}{0.5\textwidth}
% % \begin{figure}
%     \centering
%     \includegraphics[width=1.0\linewidth]{figs/sdf_optimization.pdf}
%     \caption{The visualization of the optimization process. We first randomly initialize the geometry elements and then optimize the SDF by geometry constraints. The upper two examples are selected from FormalGeo-IMO~\cite{zhang2023formalgeo}, and the lower two examples are from IMO 1959 Question 5 and IMO 2021 Question 4.}
%     \label{fig:sdf_optimization}
% \end{wrapfigure}

\section{Experiments}
In this section, we first describe the implementation details, then give results of synthesized diagrams for normal high-school level and IMO level problems in Sec.~\ref{exp_visual}, next evaluate the effectiveness of GeoSDF for solving Plane Geometry Problems (PGP) in Sec.~\ref{exp_solving}, finally show analysis of our GeoSDF in the last three sections.  
 
\subsection{Implementation Details}
Our framework is implemented in Python using the PyTorch library and is structured as a user-friendly package for public release. For the optimization, we use the AdamW optimizer for a maximum of 10,000 iterations. The learning rate follows a cosine annealing schedule, decaying from an initial value of $0.1$ down to $1 \times 10^{-6}$. A synthesis is deemed successful if the total constraint loss falls below a threshold of 0.1.

In our experiments, we select FormalGeo7k~\cite{zhang2023formalgeo} as the source datasets for providing the base geometry predicates as the input constraints of GeoSDF. FormalGeo7k consists of 6,981 PGPs, which were collected and re-annotated from previous PGP datasets, including GeoQA~\cite{geoqa}, GeoQA-Plus~\cite{geoqa+}, and Geometry3K~\cite{inter-gps}. We leverage the annotated predicates from the FormalGeo7k to synthesize geometry diagrams using GeoSDF by first transforming these predicates into our structural and constraint-based representations and subsequently optimizing the SDF to generate precise and structurally valid geometry diagrams. For more challenging problems, we use the IMO problems selected by AlphaGeometry~\cite{AlphaGeometryTrinh2024}, for which we manually annotated the constraints from the problem statements. 
% All experiments were accelerated on an NVIDIA RTX 4090 GPU.
    
\begin{figure}
    \centering
    \includegraphics[width=1.0\linewidth]{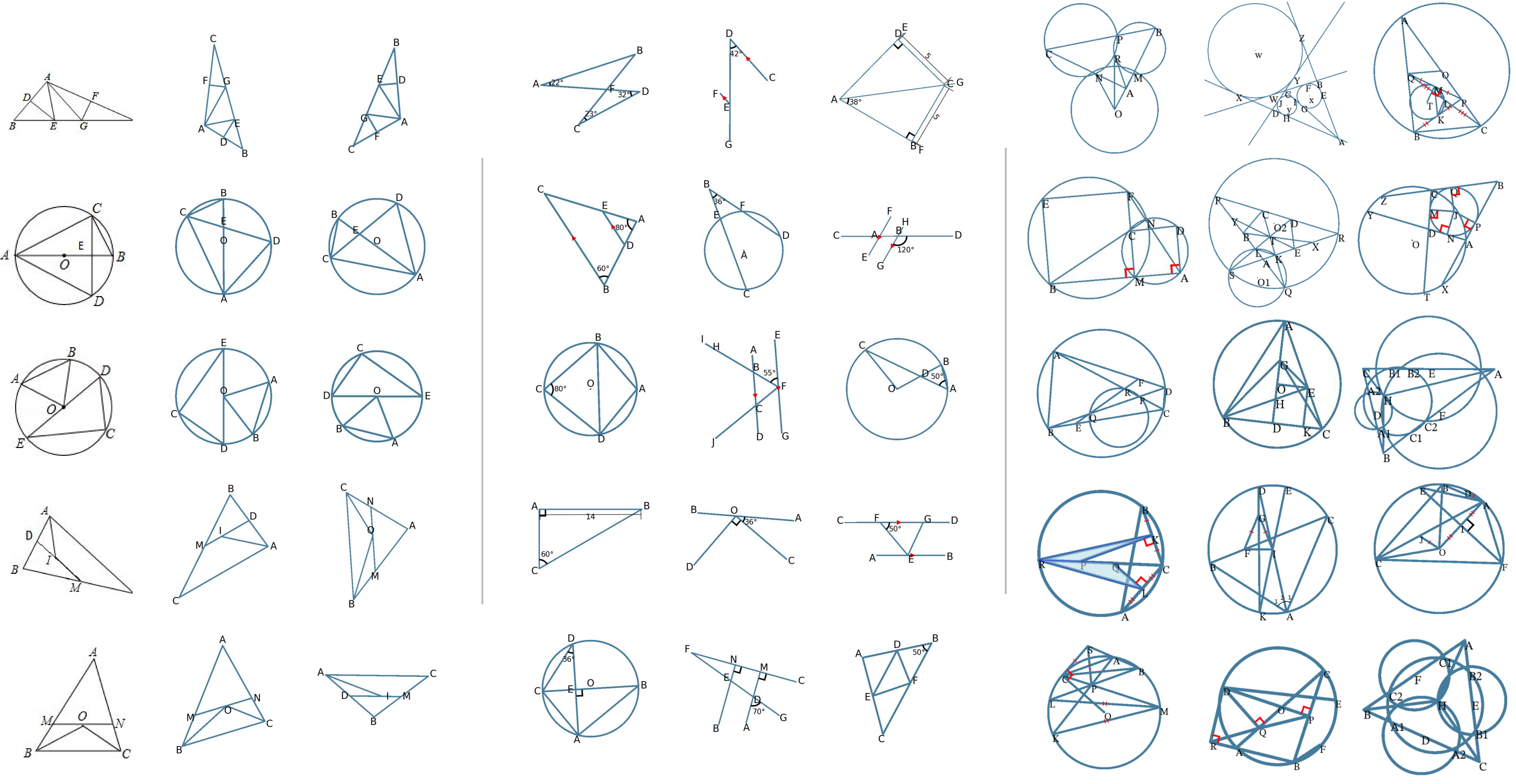}
    \caption{Qualitative results. Left: Synthesis on the FormalGeo7k dataset with diversity yet geometrically consistent (first row: ground truth). Middle: Synthesised with detailed annotations. Right: Synthesised IMO geometric problems with annotations.}
    \label{fig:sdf_visual}
\end{figure}

\subsection{Qualitative Synthesis Results}~\label{exp_visual}

    \textbf{Geometry Diagram Synthesis.}
    We first applied GeoSDF to synthesize geometry diagrams based on the FormalGeo7k dataset. For each problem in the dataset, we leveraged the annotated textual predicates as the problem constraints and then used them to synthesise the corresponding plane geometry diagram.  To demonstrate our method's ability to generate diverse outputs, we augmented the FormalGeo7k dataset by synthesizing 32 diagrams with diversity for each problem description from 7k to 224k diagrams. This process significantly increases the visual variety available for training and evaluation. A random selection of 400 synthesized examples is provided in the Appendix.

    The visualization of five cases of these synthesized diagrams is shown in the left part of Figure~\ref{fig:sdf_visual}. 
    For each original problem, we select two new diagrams from the synthesized results, with the first row showing the ground truth and the other two representing the newly synthesized ones. 
    The synthesized diagrams are geometrically consistent with the originals, but with rotation, mirror, and deformation diversity. 
    Furthermore, our method provides a range of annotations for synthesized diagrams to enhance human readability, including parallelism, perpendicularity, lengths, areas, and angles. These annotations are clearly visualized in Figure~\ref{fig:sdf_visual} (Middle and Right).

    % This capability could not only enhance human readability but also Multi-Modal LLM understanding of geometric diagrams, guiding models to focus on specific geometric relationships rather than relying solely on abstract visual features, especially since visual encoders may not be optimized for these specific tasks.

    % \textbf{Comparison to Existing Methods.}
    % A key limitation of prior methods like GeomVerse~\cite{kazemi2024geomverse} and MAVIS~\cite{mavis} is their reliance on an initial base geometry (e.g., a closed shape like a rectangle or circle) from which other elements are progressively added. In contrast, GeoSDF eliminates this requirement, capable of synthesizing an entire diagram from scratch using only points and lines, without any predefined initial shape.

% \subsection{International Mathematical Olympiad Geometry}
\textbf{IMO Geometry Diagrams Synthesis.}\label{sec:exp:imo} 
    Our GeoSDF can synthesize geometrically complex diagrams, particularly those found in International Mathematical Olympiad (IMO) problems. Previous methods, limited by their reliance on rule-based algorithms, which unable to generate such diagrams. Meanwhile, the model-based methods struggle with the deterministic rules and mathematical validity, resulting in synthesizing inaccurate diagrams.
    To demonstrate our method's ability to handle these challenges, we visualize the synthesis of IMO problems (from 1959 to 2022) in Figure~\ref{fig:sdf_visual} (right). These examples showcase our framework's ability to deal with diverse and challenging constraints. The resulting diagrams are both visually coherent and mathematically precise, demonstrating the potential of GeoSDF in handling sophisticated geometry synthesis tasks.

% \subsubsection{Human Evaluation on IMO Geometry Problems}
    To verify the correctness of our synthesized IMO diagrams, we conducted a human evaluation study. We invited 20 participants with bachelor’s degrees in science to assess whether the diagrams generated by GeoSDF are geometrically equivalent to the original diagram (i.e., ground truth). A total of 30 IMO geometry problems were used in the evaluation. On average, participants judged that \textbf{88.67\%} (26.6 out of 30) of the synthesized diagrams fully reflected the geometry information in the original problem diagram. This result highlights that GeoSDF exhibits strong capability in synthesizing diagrams for IMO-level problems. The design of the questionnaire is included in Appendix~\ref{app:questionnaire}.

\subsection{Directly Generating Diagrams from Natural Language}
We assess the natural-language–to-diagram stage by evaluating the parser that converts natural problem text into symbolic constraints. 
Constraints are compared as order-independent sets, with F1 and Jaccard similarity used to measure overlap while balancing false positives and negatives~\cite{f1score}. 
Before scoring, we apply light preprocessing: standardizing text (e.g., lowercasing, sorting arguments) and tokenizing while preserving expressions such as Collinear(A,B,C). 
All evaluations are conducted on the GeoQA\cite{geoqa} test set.
The parser achieves solid overall performance with an F1 of \textbf{87.74\%} and Jaccard of \textbf{83.53\%}, showing that most constraints are captured reliably.
These results suggest that the symbolic representation produced from natural text is able to support subsequent SDF-based optimization. 
While occasional parsing errors remain, they rarely alter the overall structure of the diagram, indicating the feasibility of the pipeline.
We further provide a complete parsing example in Appendix~\ref{Natural_Language}.

% --------------------- Experiments ---------------------
\subsection{Quantifiability of GeoSDF}\label{exp_solving}

\begin{figure}[h!]
\begin{minipage}[c]{0.48\textwidth}
One of the significant properties of GeoSDF is the quantifiability. It means that all elements of the geometry diagram can be directly quantified or measured (e.g. degree of angle) and extracted (area and length will be scaled through the scale bar of each case). For example, the first problem in Fig.~\ref{fig:sdf_consistency} asks for the measure of angle CDE (marked by the red arrow). Once the SDF optimization is complete and the diagram meets the loss constraint, we extract the angle value as the solution, which is 55. Similarly, in another example from Fig.~\ref{fig:sdf_consistency}, we determine the length of line YZ, the diameter of circle M.  This capability enables us to extend its application to a broader range of experiments, such as direct problem solving and verifying whether the synthesized diagrams precisely satisfy the original problem.
\end{minipage} %
\hfill % This command adds a flexible horizontal space to push the minipages apart.
\begin{minipage}[c]{0.49\textwidth}
    \centering
        \includegraphics[scale=0.55]{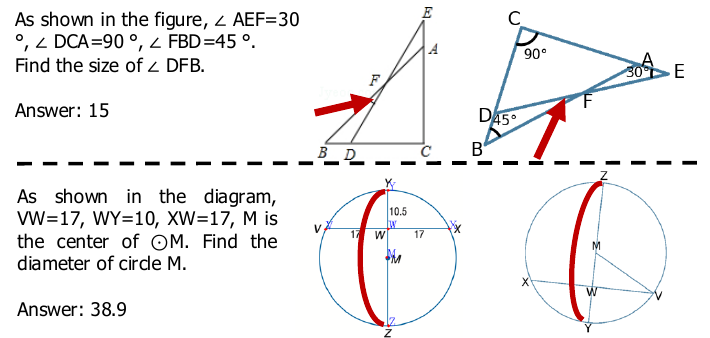}
        \caption{The example of quantifiability of GeoSDF. Our method is able to measure all the values in the diagram (e.g. degree, area), where the left diagram is the original diagram and the right one is synthesised by GeoSDF.}
        \label{fig:sdf_consistency}
\end{minipage}
\end{figure}

    \subsubsection{Plane Geometry Problem Solving}
A key application of our framework is solving Plane Geometry Problems (PGPs) through a novel "solve-by-construction" paradigm. GeoSDF finds solutions by first synthesizing a high-fidelity diagram that satisfies all given geometric constraints and then directly measuring the properties of the target element (e.g., an angle's degree or a segment's length) from the final SDF representation.

We evaluated on the GeoQA~\cite{geoqa} dataset, comparing its performance against SOTA neural and MLLM-based solvers. To provide a comprehensive analysis, we test our method in three distinct settings, with results presented in Table~\ref{tab:geosdf-solving}. GeoSDF-GT (Oracle Input): This is our main result. We use the ground-truth symbolic constraints provided in the dataset to synthesize the diagram. This setting isolates the performance of the core synthesis and measurement pipeline, removing any noise from upstream parsing errors. GeoSDF (End-to-End): To simulate a real-world scenario, this setup uses a fine-tuned LLM to parse the problem's text and diagram image into symbolic constraints, which are then fed into our solver. This evaluates the entire end-to-end pipeline.

Our primary method, GeoSDF, achieves a remarkable 94.5\% accuracy in the completion setting and 95.9\% in the choice setting. This result significantly surpasses all existing solvers, which perform logical or numerical reasoning. This highlights the potential of the solve-by-construction approach for a large part of geometry problems where the answer is a measurable quantity. Even in the end-to-end setting (GeoSDF), our method achieves SOTA accuracy of 78.5\%, demonstrating its practical viability despite the challenges of automated parsing. These experiments underscore the power of high-precision diagram synthesis as a direct path to problem-solving. Furthermore, GeoSDF's quantifiable nature makes it a promising tool for external solution verification or as a reasoning module within hybrid PGP solvers. A detailed analysis of failure cases is provided in Appendix~\ref{app:failure_analysis}.

   \begin{wraptable}{r}{0.5\textwidth}
    % \begin{table}[!h]
        \centering
            \caption{Accuracy (\%) on the GeoQA test set, the results are reported for two settings: \textbf{Completion}, the answer value is directly outputted; \textbf{Choice}, the problem choice is directly outputted. For neural PGP solvers, a random choice is selected if the completion result doesn't match any given choice.}
            \label{tab:geosdf-solving}
            \resizebox{0.5 \textwidth}{!}{
            \begin{tabular}{lcc}
                \hline
                Model                   & Completion    & Choice\\
                \hline
                \multicolumn{3}{c}{\textsc{Neural PGP Solvers}}     \\
                NGS\cite{geoqa}         &  60.0         & 69.5  \\
                DPE-NGS\cite{geoqa+}    &  62.7         & 70.4  \\
                SCA-GPS\cite{sca-gps}   &  64.1         & 72.7  \\
                \hline
                \multicolumn{3}{c}{\textsc{Closed Source MLLMs}}\\
                GPT-4o                  &  -            & 61.4  \\
                Gemini-2.5              &  -            & 63.4  \\
                \hline
                \multicolumn{3}{c}{\textsc{MLLM PGP Solvers}}    \\
                G-LLaVA-13B\cite{g-llava}& -            & 67.0   \\
                GeoX\cite{geox}         &  54.9         &  -    \\
                Qwen2.5-VL-7B\cite{qwen25vl}&  64.2     & 64.3  \\
                InternVL2.5-8B\cite{internvl}& 50.5     & 58.1  \\
                MAVIS-7B\cite{mavis}    &  -            & 68.3  \\
                R-CoT-8B\cite{r-cot}    &  -            & 75.1  \\
                GeoGen-SFT-7B\cite{pan2025enhancing} & 64.6 & 78.0 \\
                GeoUni-1.5B\cite{geouni}& 66.7          & 78.0  \\
                % NeSyGeo\cite{wu2025nesygeo}
                \hline
                % GeoSDF-PS         & 45.9 & 58.2 \\
                GeoSDF      & 78.5 & 83.2 \\
                GeoSDF-GT          & \textbf{94.5} & \textbf{95.9} \\
                \hline
            \end{tabular}}
    \end{wraptable}

    \subsubsection{Synthesised Diagram Structural Equivalence Check}
    When using GeoSDF to synthesize diagrams based on known problem statements, questions, and answers, the quantifiability of GeoSDF enables further verification of whether the synthesized diagrams satisfy the given conditions and correspond to the correct answers. If the values extracted from the SDF representation align with the expected solution and the optimization loss remains below a predefined threshold, we consider the diagram a faithful representation of the original problem. This verification process uses the problem’s goal formula to locate and extract relevant geometric features from the SDF object. To quantify this, we evaluated the success rate of diagram synthesis on the FormalGeo7K dataset. Among 6,981 samples, 5,943 diagrams (\textbf{85.13\%}) passed the loss constraint check, and 5,697 diagrams (\textbf{81.60\%}) successfully meet the target equivalence conditions. 
    Upon evaluating the dataset, we identified that the main source of synthesis failures involves problems that are not strictly value-based. These cases typically include undefined variables within their geometric constraints, such as defining a line segment's length as 'x+2'. While our parsing module capably translates these statements into symbolic constraints, the GeoSDF synthesis process cannot render a diagram without concrete numerical values, leading to a failure in visualization.

\subsection{Equivalence of Synthesised GeoSDF Diagrams}
    % \begin{table}[!h]
    \begin{wraptable}{r}{0.5\textwidth}
        \centering
        \caption{Comparison (\%) on the GeoQA test set between original diagrams and those synthesized by GeoSDF, evaluated across different base models.}
        \resizebox{0.48 \textwidth}{!}{
        \label{tab:geoqa}
            \begin{tabular}{lcc}
                \hline
                Test Set                & \makecell{Original\\Test Set}  & \makecell{Synthesised\\by GeoSDF}  \\
                \hline
                NGS~\cite{geoqa}        & 60.9          & 61.5 \\
                SCA-GPS\cite{sca-gps}   & 64.1          & 64.9 \\
                G-LLaVA-7B\cite{g-llava}& 64.2          & 64.3 \\
                \hline
            \end{tabular}
            }
    % \end{table}
    \end{wraptable}

    To further assess whether the diagrams synthesized by GeoSDF preserve the same geometry information in the original diagrams, we conducted an experiment by replacing the geometry diagrams in the test set of GeoQA with newly synthesized versions. We then evaluated whether baseline models could still solve the problems using these new diagrams. We tested several baseline models, including NGS~\cite{geoqa} and G-LLaVA-7B~\cite{g-llava}, using both the original and GeoSDF-synthesized test sets. All models were evaluated with their publicly released code, and model weights were trained on the original training set. To create the new diagrams, we annotated problem predicates and used GeoSDF to synthesize diagrams for each test example. The results of this experiment are shown in Table~\ref{tab:geoqa}. Interestingly, the accuracy on the GeoSDF-synthesized test set was slightly higher than on the original test set. We attribute this improvement to the fact that some of the original diagrams were of relatively low quality, which likely hindered model performance. In contrast, the diagrams synthesized by GeoSDF offer clearer visual representations, helping the models better understand the geometry and ultimately improving accuracy.

\subsection{Computational Efficiency and Convergence Analysis} 
    
\begin{wraptable}{r}{0.5\textwidth}
    % \begin{table}[h!]
    \centering
    \caption{Performance, Resource Usage, and Time vs. Batch Size over IMO Diagram Synthesis. The Successful Rate is calculated as the loss less than 0.1 in a batch. The Memory is the maximum GPU memory allocated throughout the entire process. We take 10000 steps for optimization.}
    \label{tab:performance}
    \resizebox{0.5 \textwidth}{!}{
    \begin{tabular}{@{}rrrrr@{}}
    \toprule
    Batch Size & Successful Rate (\%) & Memory (GB) &  Time (s) \\
    \midrule
    2048       & 57.67         & 14.49             & 98.7                \\
    1024       & 57.42         & 7.02              & 60.5                \\
    512        & 56.84         & 3.52              & 42.7                \\
    256        & 60.16         & 1.77              & 33.6                \\
    128        & 60.94         & 0.89              & 28.8                 \\
    64         & 62.19         & 0.50              & 26.5                 \\
    32         & 53.13         & 0.38              & 25.3                 \\
    \bottomrule
    \end{tabular}}
\end{wraptable}

The computational efficiency and convergence behavior of our GeoSDF framework are critical for its practical applicability.  To evaluate computational performance, we compare the performance, resource usage, and time versus Batch Size for the IMO level figure as shown in Tab.~\ref{tab:performance}. GeoSDF could also run on the CPU or the laptop platform due to it is not sensitive to the computing power. The optimization time is nearly constant versus Batch Size, and we find the bottleneck is in the transfer speed between the GPU and CPU. We measure the average optimization time for generating diverse geometric figures. For typical complex IMO level figures (e.g., those involving 20+ elements and 20+ constraints), the generation process converges in about 20s with annotations. For example, the optimization time is about 21.3s for IMO 2021 P4 for all batch size configurations with 24 elements and 24 constraints. This performance demonstrates the feasibility of our approach for interactive or near-interactive geometric design.

\subsection{Impact of Crowd Regularization} 

We conducted an ablation study to show the contribution of the crowd regularization term (i.e., Eqn.~\ref{eqn:reg}) in GeoSDF. This term aims to prevent geometric elements from collapsing, encouraging well-distributed diagrams. Without the crowd term, diagrams generated on the GeoQA dataset may degrade, with points overlapping and lines merging. Quantitatively, this resulted in a decrease of \textbf{ 3.7\%} in the success rate of diagram synthesis on the FormalGeo7K dataset, ranging from 77.89\% to 81.60\%. This demonstrates that the crowd term enforces geometric separation and visual clarity.

% \subsection{Failure Analysis} 

% \textbf{Optimization.} Since it is a non-convex optimization problem, there exist failure cases. We found that the loss threshold could be an easy and reasonable filter to drop the failure cases in an optimization batch. In the implementation, we only render the diagrams which loss below the 0.1.

% \textbf{Natural Language Parsing.} We found the model can only extract the elements, constraints, and goal, but if there needs to add some points not mentioned, the accuracy will be dropped. If the user inputs elements or constraints syntax error, the Python compiler will provide human-friendly exceptions. If the user input constraints have some conflicts, the loss of the conflicting constraints will be large. Use could know the conflicting constraints from the corresponding loss.

\section{Conclusion}
In this work, we introduced GeoSDF, a novel approach for synthesizing plane geometry diagrams by leveraging Signed Distance Field. GeoSDF can generate precise and semantically aligned diagrams, supported by an integrated self-verification mechanism that ensures consistency with the original problem constraints. By leveraging the implicit representation and differentiability of SDFs, GeoSDF enables flexible and accurate diagram synthesis. Our extensive experiments demonstrate that GeoSDF can produce high-quality diagrams, preform well across a wide range of problems, including complex International Mathematical Olympiad (IMO) problems, and effectively handle intricate geometric relationships and constraints. A key advantage of GeoSDF is its quantifiability, which opens up promising applications such as directly solving geometry problems. Experimental results show that GeoSDF significantly outperforms existing SOTA solvers, highlighting its strong potential to contribute to the field of solving geometry problems.  Meanwhile, GeoSDF also maintains strong computational efficiency, making it well-suited for deployment across various application domains with minimal computational resource requirements. Our GeoSDF bridges the gap between computational graphics and AI4Math, offering a robust tool for advancing geometric reasoning in MLLMs and beyond. Future work will explore GeoSDF to handle more types of geometry diagrams, including solid geometry and analytic geometry.

\clearpage

\section{Ethics Statement}

We have read and adhered to the ICLR Code of Ethics. Our research focuses on the automated synthesis of mathematical diagrams, a fundamental task in computational geometry and AI for education. We have considered the ethical implications of our work and foresee no direct negative societal consequences.

The datasets used in our research, FormalGeo7k and the IMO problem set, are established public benchmarks in the academic community. Our contribution includes a new dataset of synthesized diagrams, which is derived from these public sources and contains no personally identifiable or sensitive information. We believe that our tool, GeoSDF, serves a beneficial purpose by providing a means to generate high-quality educational materials and to advance research in automated mathematical reasoning.

\section{Reproducibility Statement}

To ensure the reproducibility of our work, we commit to making our code, data, and experimental setup fully available.

Code and Tool: The complete source code for the GeoSDF framework, along with the scripts required to reproduce all experimental results presented in this paper, will be made publicly available in a GitHub repository after peer review. We will also release the user-friendly online tool mentioned in the abstract.

Datasets: The experiments are based on the publicly available FormalGeo7k and IMO datasets. Furthermore, we will release the complete set of diagrams synthesized by GeoSDF for our experiments, allowing for direct verification of our results and reuse by the community.

Experimental Details: A detailed description of the implementation, hyperparameters, and optimization settings can be found in Section 4.1. Further analysis, including the failure analysis and additional results that provide deeper insight into the method's behavior, are located in the Appendix.

\bibliography{iclr2026_conference}
\bibliographystyle{iclr2026_conference}

\clearpage

\appendix

\section{Constraints Details in Mathematic}\label{app:constraints}

\textbf{Equal.} Computes the absolute difference between two values, $A$ and $B$:

\[
\text{Equal}(A,B) = |A - B|
\]

\textbf{Less.} A differentiable version of the less-than-or-equal constraint. 
\[
\text{less}(A, B) = \max(A - B, 0)
\]
This can also be written as:
$$
\text{less}(a, b) = 
\begin{cases} 0 & \text{if } a \leq b \\ a - b & \text{if } a > b  \end{cases}
$$

\textbf{Angle.} For three points \(A\), \(B\), and \(C\) (with \(B\) as the vertex), the angle \(\theta\) at \(B\) is computed using the dot product of vectors \(\vec{BA}\) and \(\vec{BC}\):
\[
\text{Angle}(A, B, C) = \arccos\!\left(\frac{\vec{BA} \cdot \vec{BC}}{\|\vec{BA}\|\|\vec{BC}\|}\right),
\]
where $\vec{BA} = A - B,\vec{BC} = C - B$. 

\textbf{Area.} 
Given vertices \((x_1, y_1), (x_2, y_2), \dots, (x_n, y_n)\) of a polygon (assuming \((x_{n+1}, y_{n+1}) = (x_1, y_1)\)), the area is:
\[
\text{Area}((x_1, y_1), \dots, (x_n, y_n)) = \frac{1}{2}\left|\sum_{i=1}^{n}\left(x_i y_{i+1} - y_i x_{i+1}\right)\right|
\]

\textbf{Parallel.} For two line segments \(AB\) and \(CD\), let their direction vectors be
\[
v_1 = B - A,\quad v_2 = D - C.
\]
In 2D, the cross product (a scalar) that measures the non-parallelism is:
\[
L_{\text{parallel}} = \left| v_{1x}v_{2y} - v_{1y}v_{2x} \right|
\]
Next, define the minimum distance between the segments as:
\[
d = \min\{d(A, \overline{CD}),\; d(B, \overline{CD}),\; d(C, \overline{AB}),\; d(D, \overline{AB})\}
\]
A distance penalty is applied if this distance is below a threshold \(T\) (with weight \(\lambda\)) to avoid the elements crowd:
\[
P_{\text{distance}} = \lambda\,\max\{0,\, T - d\}
\]
Thus, the final constraint is:
\[
\text{Parallel}(AB,CD)= \left| v_{1x}v_{2y} - v_{1y}v_{2x} \right| + \lambda\,\max\{0,\, T - d\}
\]

\textbf{Perpendicular.} When the lines are defined by endpoints \(A_1A_2\) and \(B_1B_2\), let:
\[
v_1 = A_2 - A_1,\quad v_2 = B_2 - B_1.
\]
The perpendicularity constraint remains:
\[
\text{Perpendicular}(AB,CD) = \left| v_1 \cdot v_2 \right|
\]

\textbf{Order.} For points on a circle with center \(O\), let the angle from \(O\) to each point be \(\theta_i\) (after unwrapping so they are continuous). The difference between consecutive angles is:
\[
\Delta \theta_i = \theta_{i+1} - \theta_i
\]
A penalty is applied when a difference is too small (below a small constant \(\epsilon\)):
\[
\text{Order}((x_1, y_1), \dots, (x_n, y_n)) = \sum_{i=1}^{n-1} \max\{0, \epsilon - \Delta \theta_i\}
\]
\textbf{Inside.} Determines whether a point lies inside a convex polygon. Given a point $P$ and a convex polygon defined by vertices $V_1, V_2, \ldots, V_n$ ordered either clockwise or counterclockwise. For each edge of the polygon, we define vectors:
$$
\vec{e}i = \vec{V}{i+1} - \vec{V}_i \quad \text{for } i = 1,2,\ldots,n,
$$
where $\vec{V}_{n+1} = \vec{V}_1$ (to close the polygon) For each vertex, we compute vectors from the vertex to point $P$:
$$
\vec{d}_i = \vec{P} - \vec{V}_i \quad \text{for } i = 1,2,\ldots,n,
$$
Calculate the cross products between edge vectors and vectors to point $P$:
$$
c_i = \vec{e}i \times \vec{d}i = (e{i,x} \cdot d{i,y} - e_{i,y} \cdot d_{i,x}) \quad \text{for } i = 1,2,\ldots,n.
$$
For a point inside the polygon, all cross products should have the same sign. Let $s_1 = \text{sign}(c_1)$ be the sign of the first cross product. The penalty for each edge is defined as:
$$
p_i = \max(0, -s_1 \cdot c_i) \quad \text{for } i = 1,2,\ldots,n.
$$
The total penalty is the sum of all individual penalties:
$$
\text{penalty} = \sum_{i=1}^{n} p_i
$$
If $\text{penalty} = 0$, then point $P$ is inside the convex polygon. Otherwise, $P$ is outside the polygon.

\textbf{Crowd Penalty.} Given \(N\) points \(x_1, x_2, \dots, x_N \in \mathbb{R}^D\), define the Euclidean distance between points \(x_i\) and \(x_j\) as:
\[
d_{ij} = \|x_i - x_j\|
\]
The penalty for each pair (only considering \(i < j\)) is:
\[
p_{ij} = \max\{0, T - d_{ij}\}^2
\]
The Crowd Penalty is:
\[
P_\text{Crowd} = \sum_{1 \le i < j \le N} \max\{0, T - d_{ij}\}^2
\]

We make this algorithm vectorized to speed up the calculation in GPU.

\clearpage

\section{Optimization Process Visualization}\label{app:optimization_process_visualization}

\begin{figure}[h]
    \centering
    \includegraphics[width=1.0\linewidth]{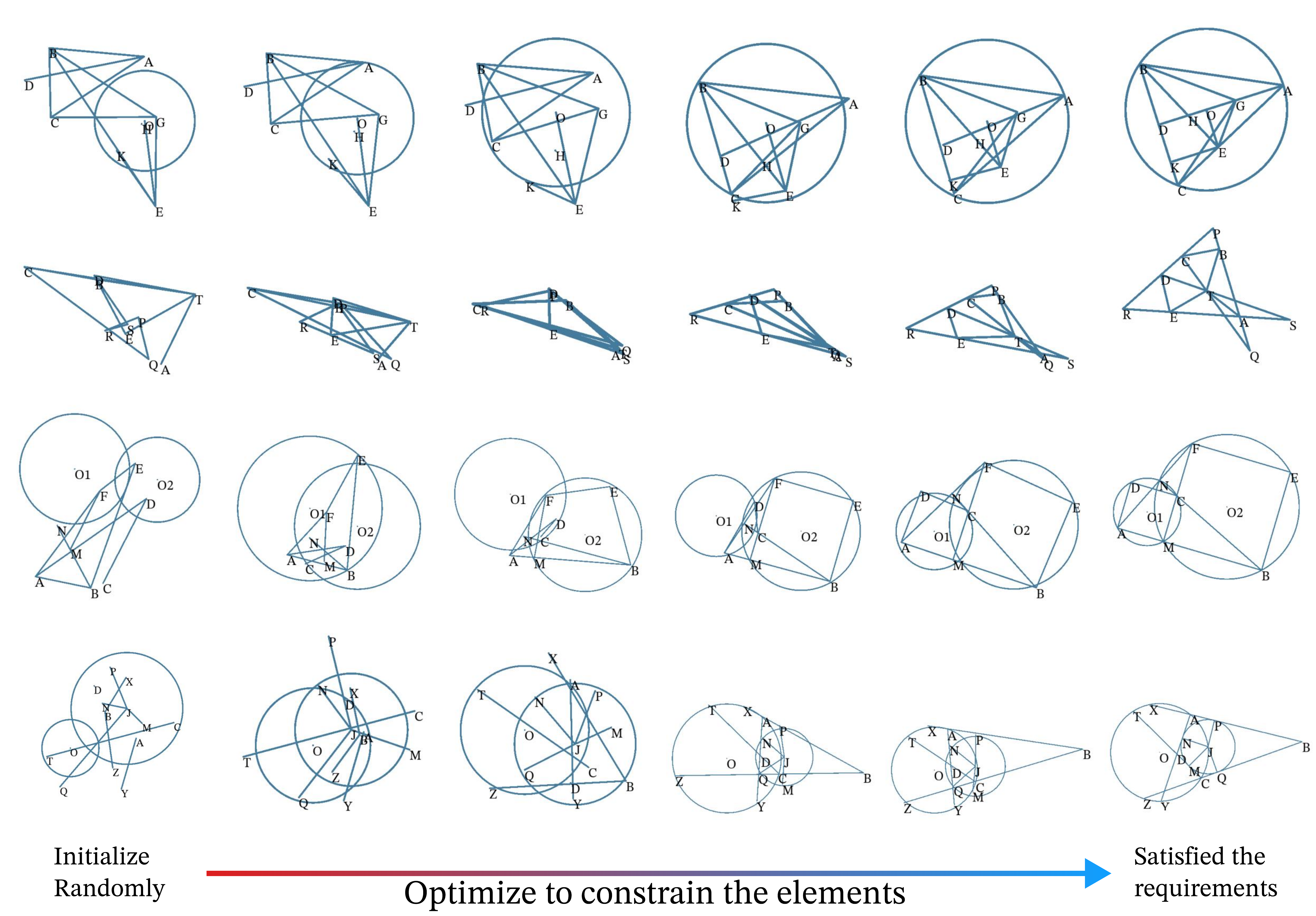}
    \caption{The visualization of the optimization process. We first randomly initialize the geometry elements and then optimize the SDF by geometry constraints to the synthesized geometry diagram. The upper two examples are selected from FormalGeo-IMO~\cite{zhang2023formalgeo}, and the lower two examples are from IMO 1959 Question 5 and IMO 2021 Question 4.}
    \label{fig:sdf_optimization}
\end{figure}

We visualized the process of the optimization of the GeoSDF in Fig.~\ref{fig:sdf_optimization} with four IMO examples of the geometry diagrams. The leftmost diagrams represent the initial state, while the rightmost diagrams depict the final state after the optimization process. The initial state is set randomly. Through optimization, the SDF diagrams gradually evolve from a random configuration \( E \) to \( E^* \), ultimately satisfying the diagram's constraints. 

\clearpage

\section{Boundary Threshold}\label{app:boundary_threshold}

\begin{figure}[ht]
    \centering
    \includegraphics[width=1.0\linewidth]{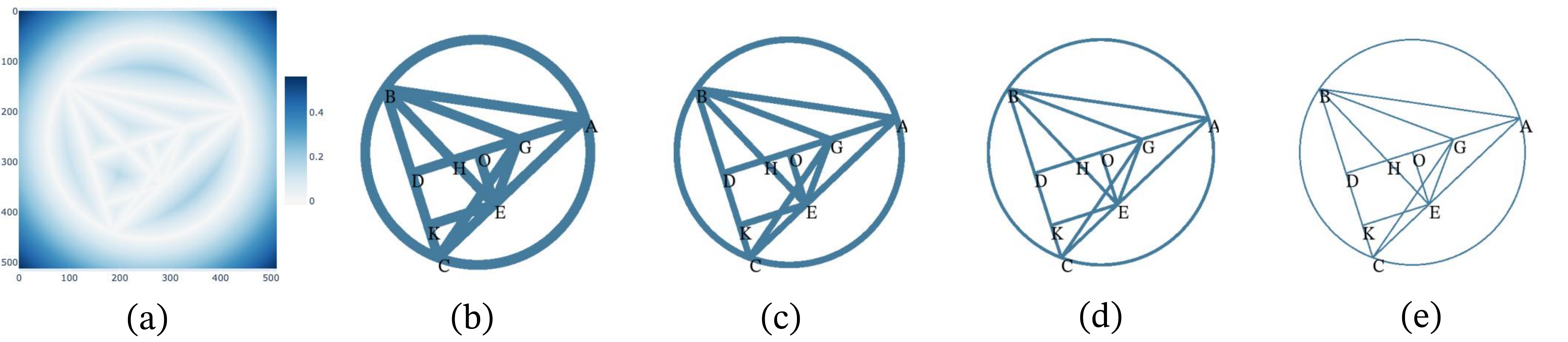}
    \caption{(a) The values sampled from SDF by $512\times512$ pixel grid visualization. And visualizations of its zero iso-surface with different thresholds (b) 0.3, (c) 0.2, (d) 0.1, (e) 0.005.}
    \label{fig:threshold}
\end{figure}

The optimized configuration $E^*$ allows for sampling the SDF of the geometric elements and evaluating the zero iso-surface to render the corresponding point set. To achieve this, a $N^2$ pixel grid is sampled. A threshold $\tau$ is then applied to the SDF values to extract the boundaries of the geometric shapes. This threshold determines the sampling range for the field, and its impact on the visualization is presented in Figure \ref{fig:threshold}. Different threshold values result in varying levels of detail in the geometric diagram; specifically, a higher threshold expands the element boundaries, leading to thicker lines.

\clearpage

\section{Failure Analysis}\label{app:failure_analysis}

While GeoSDF demonstrates strong performance, it is important to analyze its failure modes, which primarily fall into two categories: optimization challenges and parsing limitations.

\textbf{Optimization Failures.} The core of our method relies on solving a non-convex optimization problem. Consequently, the optimizer can sometimes converge to a local minimum or a saddle point that does not represent a valid geometric configuration. These failures manifest in two ways mainly:

(a) High Final Loss: This failure case occurs when the optimization process terminates without satisfying all constraints, resulting in a high total loss. This happens in highly constrained problems where the optimization is particularly complex. We treat these cases as synthesis failures. To deal with this type of failure, a convergence threshold is set to filter the failure cases out. In the implementation, all the diagrams that loss above our convergence threshold of $0.1$ are dropped out.

(b) Geometric Degeneracies: In some instances, the optimizer might produce a diagram with a low loss, but which is still visually or geometrically incorrect. This can include degenerate cases like three points becoming collinear when they should form a non-trivial triangle, or elements overlapping in a way that violates implicit assumptions of the problem. Our crowd regularization term (Eqn.~\ref{eqn:reg}) mitigates many of these issues, but they are not entirely eliminated.

\textbf{Natural Language Parsing Limitations.} The accuracy of the entire pipeline is fundamentally dependent on the initial NL-to-symbolic parsing step. We identified two primary sources of error from our fine-tuned LLM parser:

(a) Implicit Information: Many geometry problems contain implicit information that is not explicitly stated in the text. For example, a problem might refer to the intersection of two lines without explicitly defining a point at that intersection. Our current parser struggles to infer and create these necessary-but-unstated geometric elements, leading to an incomplete set of constraints for synthesis.

(b) Ambiguity and Errors: Ambiguous phrasing in the problem description can lead to incorrect symbolic translations. Furthermore, while our framework can detect syntactical errors (via the Python compiler) and logically conflicting constraints (via high residual loss on the conflicting terms), it cannot correct semantic mistakes by the user. For example, if the user misinterprets "tangent to" as "intersecting," the resulting diagram will be fundamentally flawed despite the optimization succeeding on the misinterpreted constraints. Diagnosing these failures is straightforward. By inspecting the per-constraint loss values, users can immediately identify which geometric relationships were not satisfied, providing clear feedback on the source of the optimization failure.

\clearpage

\section{Examples of IMO Figure Synthesize Process}\label{app:Synthesize_Process}

In this section, we present a complete IMO geometric synthesis process implemented in Python. It should be noted that the initialization of elements and constraints can be executed automatically. These elements are maintained here for completeness.

\begin{figure}[h!]
    \begin{minipage}[c]{0.5\textwidth}
        
\begin{small}
\begin{verbatim}
from sdf2d import *
# the set of elements E
# batch size is 64
s = Shape(batch=64) 
s.A = Point()
s.B = Point()
s.C = Point()
s.D = Point()
s.E = Point()
s.G = Point()
s.H = Point()
s.K = Point()
s.O = Point()
s.AB = Segment(s.A, s.B)
s.AC = Segment(s.A, s.C)
s.AD = Segment(s.A, s.D)
s.BC = Segment(s.B, s.C)
s.BE = Segment(s.B, s.E)
s.BG = Segment(s.B, s.G)
s.CG = Segment(s.C, s.G)
s.EG = Segment(s.E, s.G)
s.EK = Segment(s.E, s.K)
s.EO = Segment(s.E, s.O)
s.circle_O = Circle(center=s.O) 
s.render() # sample the SDF
\end{verbatim}
\end{small}
    \end{minipage} %
    \begin{minipage}[c]{0.5\textwidth}
        \centering
    \includegraphics[width=1\linewidth]{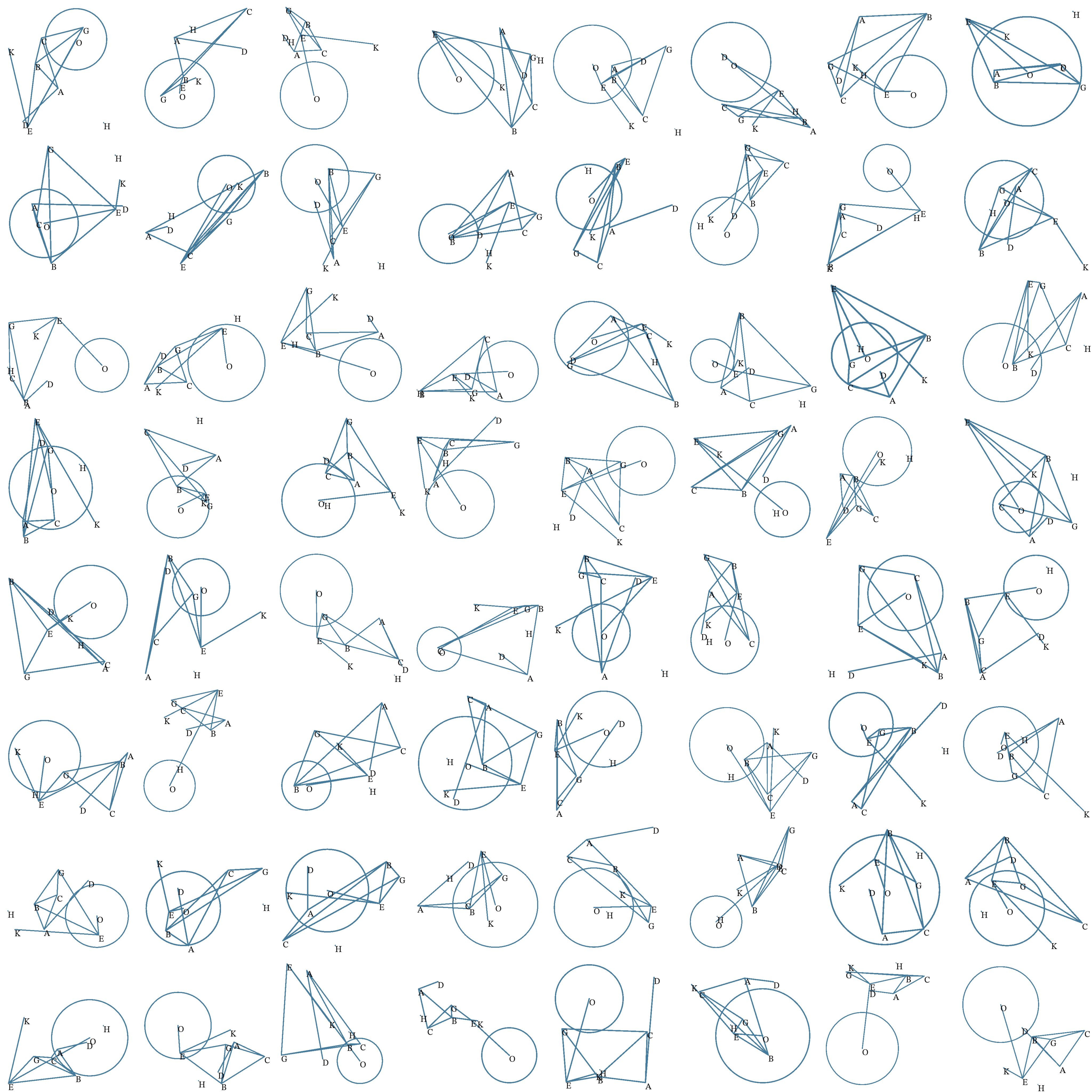}
    \caption{Initial state diagram in a batch, containing 64 samples.}
    \label{fig:initial_1}
    \end{minipage}
\end{figure}

\begin{figure}[h!]
    \begin{minipage}[c]{0.5\textwidth}

\begin{small}
\begin{verbatim}
def constraints():
    return (
        s.AD(s.G)
        + s.AD(s.O)
        + s.AD(s.H)
        + s.BE(s.H)
        + s.BC(s.D)
        + s.BC(s.K)
        + s.AC(s.E)
        + s.circle_O(s.A)
        + s.circle_O(s.B)
        + s.circle_O(s.C)
        + equal(s.A - s.B, s.A - s.C)
        + equal(s.A - s.H,  
          2*(s.D - s.circle_O))
        + equal(s.A - s.G, s.H - s.G)
        + parallel(s.EO, s.BC)
        + perpendicular(s.BE, s.AC)
        + perpendicular(s.BC, s.EK)
        + crowd_penalty(s.points())
    )
s.optimize(constraints)
s.render()

\end{verbatim}
\end{small}
\end{minipage} %
\begin{minipage}[c]{0.5\textwidth}
    \centering
\includegraphics[width=1\linewidth]{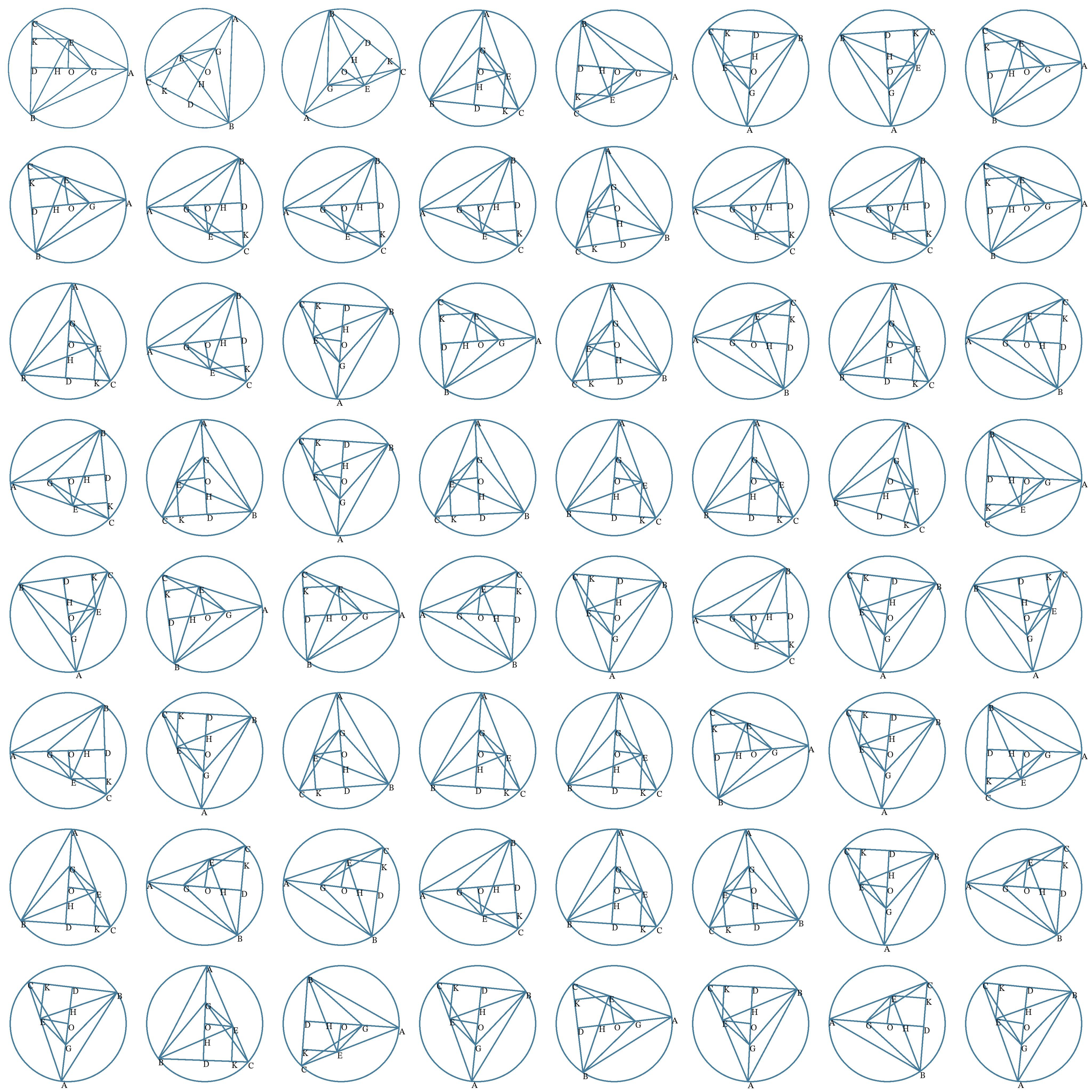}
\caption{Batched results in final state after optimization.}
\label{fig:final_state_1}
\end{minipage}
\end{figure}

The initial state figure will be synthesized as Fig.~\ref{fig:initial_1} with random parameters for each element. Then, define the constraints based on the mathematical geometric conditions. Finally, optimize the constraints and render the Fig.~\ref{fig:final_state_1}.

\clearpage

\section{Synthesis Geometric Diagram from Natural Language.}\label{Natural_Language}

\begin{figure}[h]
    \centering
    \includegraphics[width=0.65\linewidth]{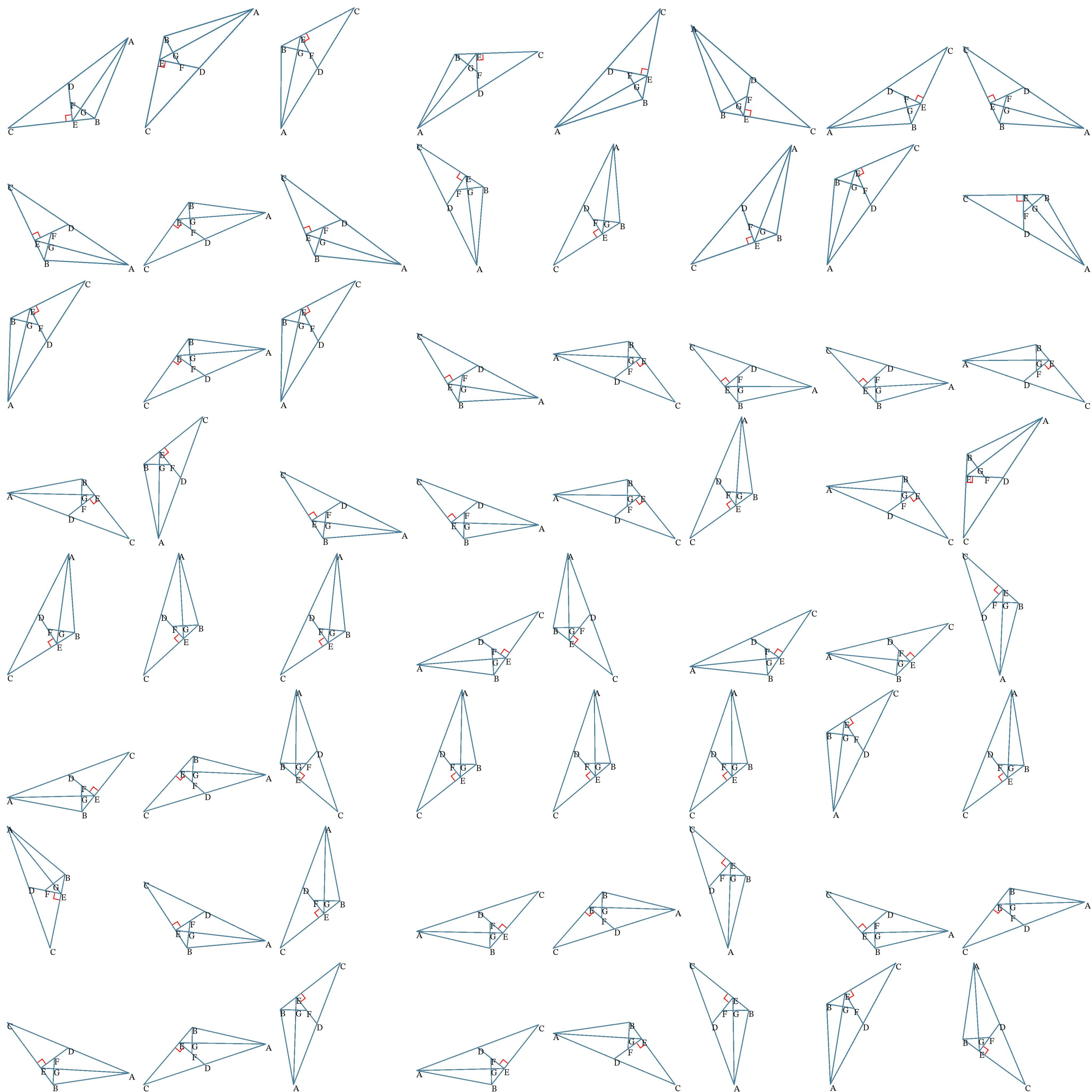}
    \caption{A batch of synthesised output diagrams by using natural language.}
    \label{fig:language_out}
\end{figure}

Here is an example of using a fine-tuned LLM to convert the natural language to the Diagram. The synthesized diagram is shown as Figure~\ref{fig:language_out}.

\textbf{Prompt: }Convert the given problem text into the Symbolic Language of Elements, Constraints, and the Goal. Problem Text: In the triangle ABC, AB=BC, D is the midpoint of AC. DE $\perp$ BC on E. The F is the midpoint of DE. G is the intersection point of AE and BF. Find the angle of AGB.

\textbf{Output: }

\begin{figure}[h!]
    \begin{minipage}[c]{0.4\textwidth}
        
\begin{small}
\begin{verbatim}
### Elements: 
s = Shape(B=64)
s.A = Point()
s.B = Point()
s.C = Point()
s.D = Point()
s.E = Point()
s.F = Point()
s.G = Point()

s.AB = Segment(s.A, s.B)
s.BC = Segment(s.B, s.C)
s.AC = Segment(s.A, s.C)
s.AE = Segment(s.A, s.E)
s.BF = Segment(s.B, s.F)
s.DE = Segment(s.D, s.E)
\end{verbatim}
\end{small}

\end{minipage} %
\begin{minipage}[c]{0.5\textwidth}
\begin{small}
\begin{verbatim}
### Constraints: 
def constraints_fn(step):
    return (
        +s.BC(s.E)
        + s.AC(s.D)
        + s.AE(s.G)
        + s.BF(s.G)
        + s.DE(s.F)
        + eq(s.A - s.B, s.B - s.C)
        + eq(s.A - s.D, s.D - s.C)
        + eq(s.D - s.F, s.E - s.F)
        + perpendicular(s.DE, s.BC)
        + crowd_penalty(s.points_tensor())
    )

### Goal: 
angle(s.A, s.G, s.B)
\end{verbatim}
\end{small}
\end{minipage}
\end{figure}

\clearpage
\section{The Design of the Questionnaire.}\label{app:questionnaire}
\begin{figure}[h]
    \centering
    \includegraphics[width=0.8\linewidth]{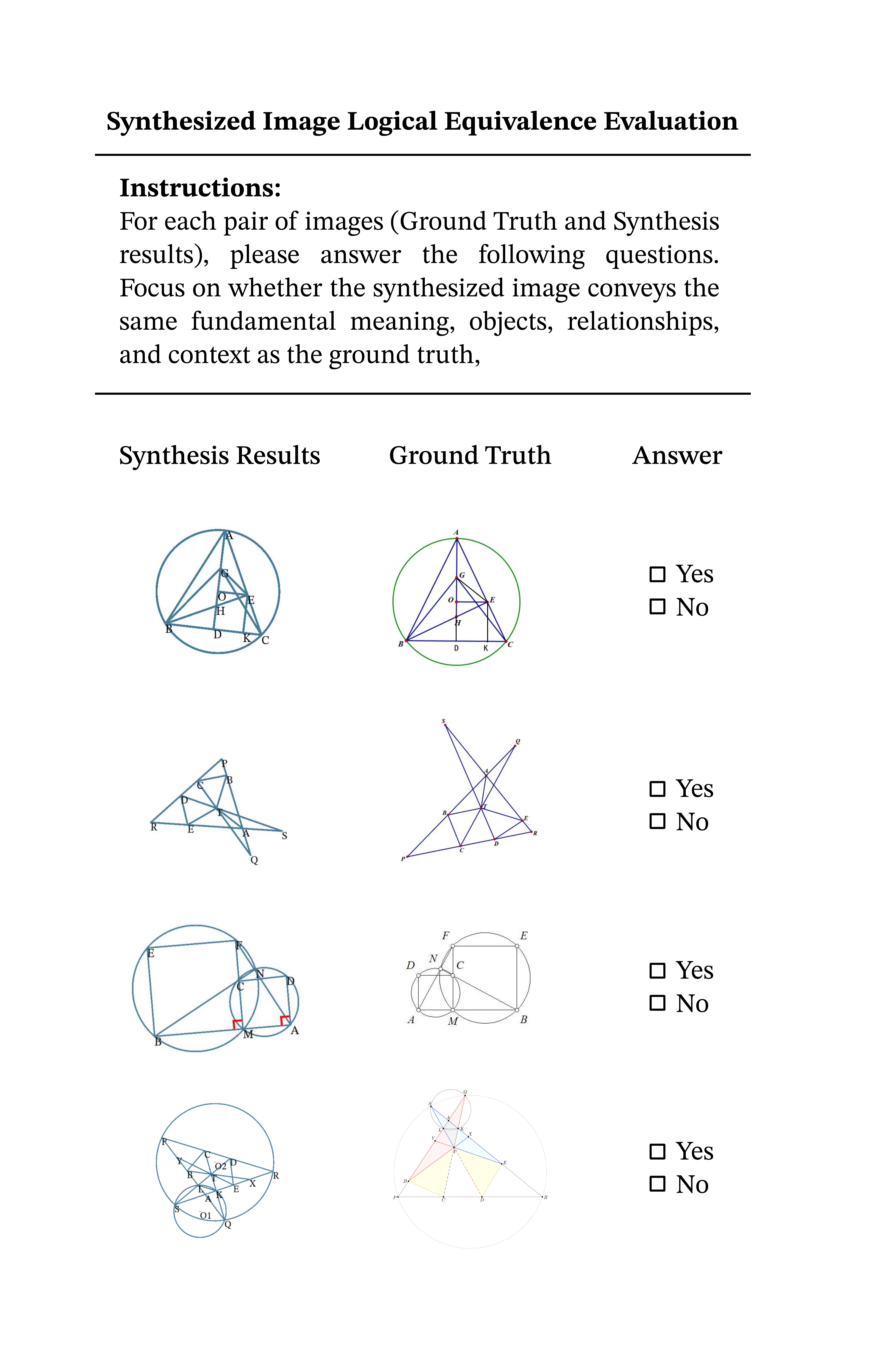}
    \caption{The design of the questionnaire.}\label{fig:questionnaire}
\end{figure}

The Figure~\ref{fig:questionnaire} is an example we designed for the questionnaire. 

\clearpage

\section{Signed Distance Field}\label{app:sdf}
\textbf{Signed Distance Field Definition} A field in mathematics is a function that assigns a value (scalar or vector) to each point in a space. The space provides the coordinate system and the notion of distance necessary. The Signed Distance Field assigns a signed distance to the nearest point on the boundary of the shape for every point in this space. In mathematics, the SDF \(F\) is a scalar field that assigns a value $s$ to every position $x$ in the space by 
$$F(x) = s, \quad x\in \mathbb{R}^2,\, s\in \mathbb{R},$$ 

where the value $s$ indicates the distance from the position $x$ to the shape \(\Omega \in \mathbb{R}^2\) in the two-dimensional plane space. The distance between a point \(x\) and the boundary of the shape \(B\) is hereby defined as
\begin{equation}
d(x, B) = \inf\limits_{y\in B} d(x, y),
\end{equation}
where \( \inf \) denotes the infimum and \( d \) represents the distance. For simplicity, that means the assigned value to the space corresponds to the minimum distance from the point to the geometric shape.

\textbf{SDF Related Work.} SDFs have been widely explored for their versatility in representing and rendering complex 3D geometries. Early foundational work by Frisken et al.~\cite{frisken2000adf} introduced Adaptively Sampled Distance Fields (ADFs), optimizing memory usage while preserving geometric detail. SDFs have since been instrumental in real-time rendering, with Quilez~\cite{quilez2008raymarching} popularizing their use in procedural ray marching and Green~\cite{green2007sdftext} introducing their application for scalable and anti-aliased text rendering. Advances in GPU optimization, as explored by McGuire et al.~\cite{mcguire2013gpu}, have further enhanced the real-time capabilities of SDFs. Techniques like the Marching Cubes algorithm~\cite{lorensen1987marching} and Dual Contouring~\cite{ju2002dual} enable the conversion of volumetric SDFs into detailed meshes, critical for physics simulations and surface reconstruction. In physics-based applications, Fournier and Rico~\cite{fournier2006collision} leveraged SDFs for efficient collision detection and soft body interactions. More recently, neural approaches like DeepSDF~\cite{park2019deepsdf} and NeRF-SDF~\cite{yariv2020nerfsdf} have demonstrated the power of combining SDFs with deep learning for shape representation and neural rendering. Furthermore, sparse data structures, such as OpenVDB~\cite{museth2013openvdb}, address memory and computation challenges in large-scale volumetric datasets. These works underscore the importance of SDFs in advancing real-time rendering, procedural modeling, and computational efficiency, forming the foundation upon which this research builds. In our work, we highlight the differentiable property of the SDF to generate math figures using math conditions. 

\section{Interactive Online Tool}
To promote broader use and facilitate further research, we have developed an interactive online tool based on GeoSDF. This tool enables the community to synthesize geometric diagrams by providing input as either natural language text or Python code. It serves as a practical demonstration of our framework and a valuable resource for researchers and educators. A demonstration video showcasing the tool's functionality is included in the supplementary materials.
\clearpage

\section{IMO Batch Visualization}\label{app:IMO_batch_visalization}

\begin{figure*}[h!]
    \centering
    \includegraphics[width=1\linewidth]{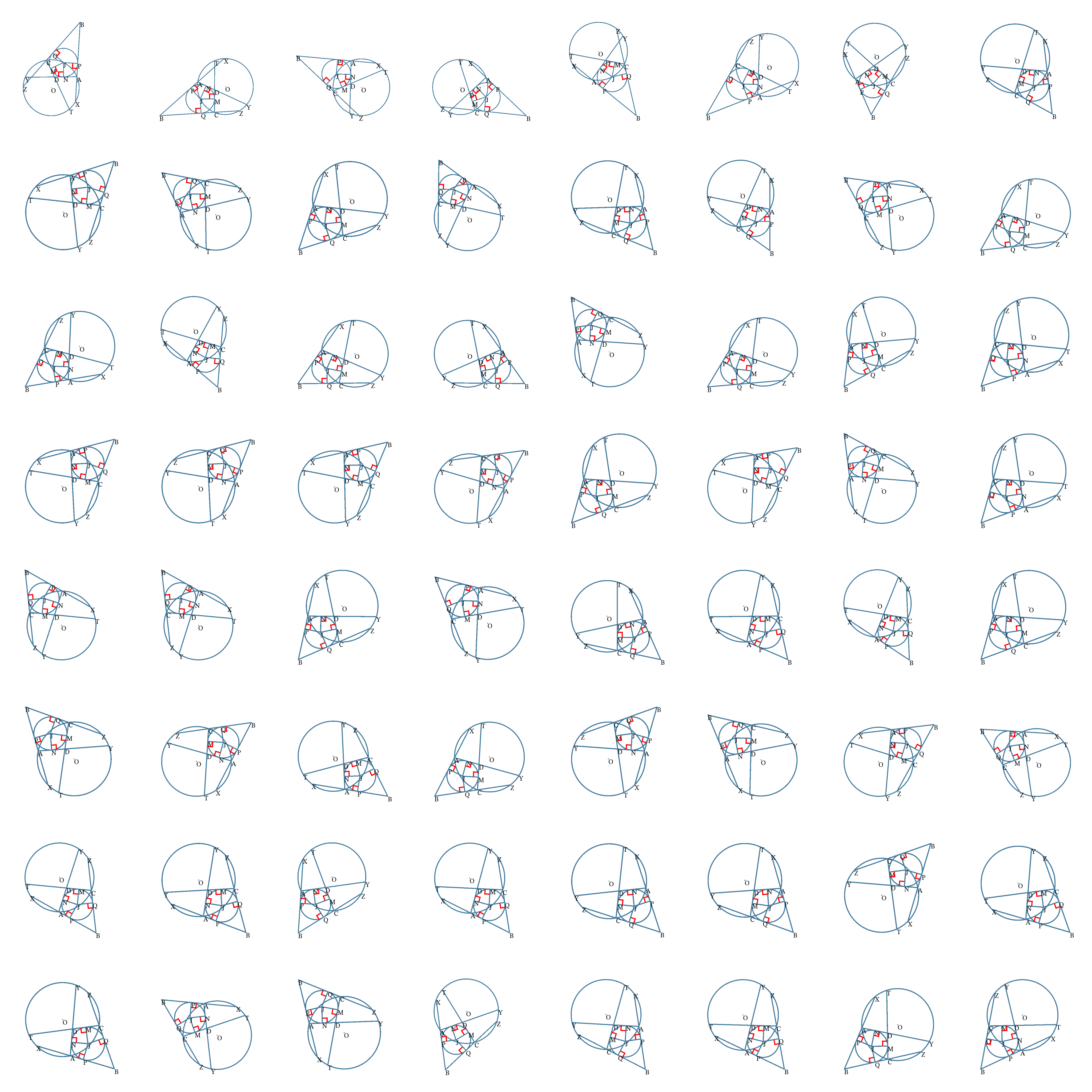}
    \caption{A batch of synthesis diagrams. Source from IMO 2021 Problem 4.}
    \label{fig:IMO_full_page_0}
\end{figure*}

\begin{figure*}[h!]
    \centering
    \includegraphics[width=1\linewidth]{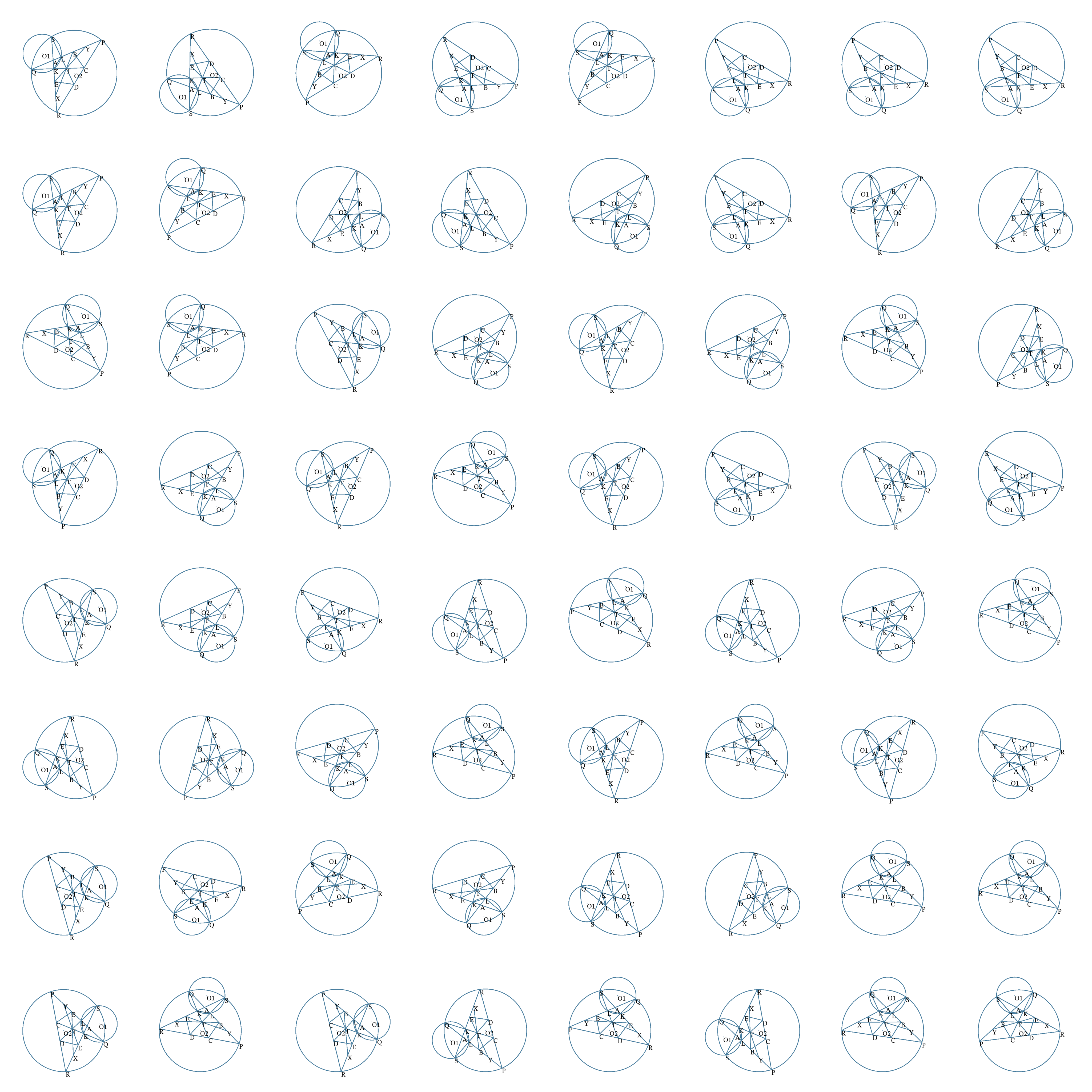}
    \caption{A batch of synthesis diagrams. Synthesis from IMO 2022 Problem 4.}
    \label{fig:IMO_full_page_1}
\end{figure*}

We show two IMO batch visualization cases. In the Fig.~\ref{fig:IMO_full_page_0} and Fig.~\ref{fig:IMO_full_page_1}, we can see the synthesis results are accurate and geometrically consistent.

\clearpage

\section{Use of Large Language Models}
In accordance with the submission guidelines, we declare that no large language models (LLMs) or other AI-powered generative tools were utilized in any phase of this work. This includes, but is not limited to, research ideation, experimental design, data analysis, and the writing and editing of this manuscript. The entirety of the work presented was conducted by the human authors.

\end{document}